\documentclass[11pt,letterpaper]{article}

\usepackage[margin=1in]{geometry}
\usepackage[utf8]{inputenc}
\usepackage[T1]{fontenc}
\usepackage{lmodern}
\usepackage{microtype}
\usepackage{amsmath}
\usepackage{amssymb}
\usepackage{amsthm}
\usepackage{graphicx}
\usepackage{booktabs}
\usepackage{array}
\usepackage{xcolor}
\usepackage[hidelinks]{hyperref}
\usepackage{enumitem}
\usepackage{authblk}
\usepackage{titlesec}
\usepackage{caption}
\usepackage{float}
\usepackage{natbib}

\titlespacing*{\section}{0pt}{1.5em}{0.6em}
\titlespacing*{\subsection}{0pt}{1.2em}{0.4em}

\newcommand{\toarrow}{$\rightarrow$}

\title{\textbf{Position Bias is Hidden Behind Ceiling Effects:\\
A Permutation Diagnostic for LLM Benchmarks}}
\author{Hiroki Tamba\thanks{Independent researcher.\\Email: \texttt{contact@tamba-research.academy}\\Code and data: \url{https://github.com/TambaClan/inspect_permute}}}
\affil{}
\date{\today}

\begin{document}

\maketitle

\begin{abstract}
Position bias in multiple-choice LLM evaluation is widely cited as a confound in capability comparisons, but published measurements rely on single answer-order shuffles whose results confound the bias signal with content-level noise and sampling stochasticity. I introduce \texttt{inspect\_permute}, an open-source extension to the inspect\_ai evaluation framework that runs exhaustive answer-order permutations per question and reports the chi-squared / Cram\'er's V signature of position bias with bootstrap confidence intervals. I apply the tool across four vendors and five MMLU subjects --- gpt-4o-mini, claude-haiku-4-5, gemini-2.5-flash, grok-3, on elementary\_mathematics, college\_mathematics, formal\_logic, professional\_medicine, and philosophy --- for a total of 24{,}000 API calls under temperature-0 generation, with falsifier predictions pre-registered via a SHA-256 hash posted publicly before half the data was observed. Position bias turns out to be statistically detectable only within a roughly 60--95\% base-accuracy Goldilocks zone. Below the zone, processing-load dominance swamps subject-specific signal. Above it, ceiling effects compress the variance below the chi-squared test's resolution. Within the zone, the detectable cells separate into two mechanism types with distinct shape signatures: monotone A\toarrow{}D decrease (\texttt{processing\_load}, observed in low-tier models across subjects) and non-monotone D-drop (\texttt{content\_ambiguity}, observed only in subject-intrinsic noisy domains and only in a narrow capability band). Standard MMLU subjects place every frontier-tier model in the sweep above the detection band, so the absence of bias signal in those cells should be read as ``not measurable on this instrument,'' not ``model is unbiased.'' Together with the ceiling-effect characterisation reported in arXiv 2606.26185, the present work brackets the detectable region of position-bias measurement from above and below, and makes the central question of the field --- whether frontier models actually carry position bias when measured by an instrument calibrated to their capability level --- askable in a verifiable form. The package and the matrix data are open under MIT, with the preregistration verifiable from a public hash anchor.

\end{abstract}

\section{Introduction}\label{sec:intro}

\subsection{Position bias as a measurement problem, not a model property}

Multiple-choice question (MCQ) benchmarks drive a lot of what we say about LLM capability. MMLU~\citep{hendrycks2021}, BBH, AGIEval, GPQA, and the safety benchmarks built on top of them are all MCQ-formatted, and a large share of the accuracy numbers cited in vendor announcements, regulatory submissions, and pre-deployment evaluations comes out of this format. The format is convenient. It turns open-ended model behaviour into a verdict against a labelled ground truth, with a clear scoring rule and minimal post-processing.

Position bias is the observation that an MCQ-format model's response can depend on the presented order of the answer options, even when the content of the options is held fixed. A model that prefers position A regardless of correctness will score higher on benchmarks whose ground-truth answers land at A more often, and lower on benchmarks where they don't. The effect is documented across multiple model families and benchmark settings~\citep{zheng2024,pezeshkpour2024,wang2024beyond} and is widely cited as a confound in capability comparisons.

But the empirical picture is unsettled. Reported magnitudes vary across studies. Reported mechanisms (first-option preference, last-option avoidance, alphabetical anchoring) often aren't distinguished from one another. And the question of which models actually carry which bias under which benchmark conditions has resisted clean answers.

A lot of this disagreement isn't about the models. It's about the measurement instrument. Most published measurements use a single answer-order shuffle per question, and the resulting estimate confounds three sources of variation: the bias signal itself, content-level noise across the eval set, and the model's residual sampling stochasticity. With $k=4$ options, a single shuffle samples one of 24 orderings; the response under the other 23 is not observed and cannot be inferred from accuracy alone. The instrument can't tell us whether a flat result reflects an unbiased model or simply an unlucky draw.

\subsection{The measurement gap}

The first gap is methodological. Exhaustive answer-order permutation --- running the same question against all $k!$ orderings and recording the response surface --- eliminates the unobserved-permutation problem, at the cost of $k!$ times the API budget. For $k=4$ that's 24 calls per question, which is manageable at the question counts typical of bias studies and trivial for vendors but historically uncommon in published protocols. The infrastructure for doing it at scale, inside an evaluation framework that already supports the rest of the pipeline (dataset loading, model adapters, result aggregation), didn't exist before this work.

The second gap is interpretive. Even when full permutation data is collected, the standard chi-squared test reports whether bias is present but not which kind. Two distinct mechanisms get hypothesised in the literature: monotone preference for early-presented positions (likely arising when the model commits to the first plausible answer under reasoning-budget pressure) and non-monotone anchoring effects (likely arising when the model has plenty of computational budget but the answer space is genuinely ambiguous). These two have different implications for benchmark design, model alignment work, and downstream interpretability claims. They share the same chi-squared signature when the test statistic is summarised as a single $p$-value.

The third gap is empirical. Even with the methodology and the interpretive vocabulary in place, position bias across model capability tiers has been studied one model at a time, with each study focusing on a single vendor or a narrow set of benchmarks. The shape of the cross-vendor matrix --- which model shows which mechanism on which subject --- has not been laid out as a single comparable object.

\subsection{What this paper contributes}

I address the three gaps in order. The contributions are presented separately so a reader can evaluate them independently.

The first is \texttt{inspect\_permute}, an open-source extension to the inspect\_ai evaluation framework~\citep{inspect_ai}. The package implements exhaustive (and sampled) answer-order permutation as a drop-in solver, records per-run position metadata via a custom scorer, and produces a chi-squared / Cram\'er's V report with bootstrap confidence intervals. It is MIT-licensed, reproducibility-tested across 31 unit tests (numerical reference plus adversarial edge cases) with 99\% line coverage, and validated against 24{,}000 real API calls during the present work. The package is available on GitHub at \texttt{TambaClan/inspect\_permute} and via \texttt{pip install inspect\_permute}. The package design and reproducibility infrastructure are described in Section~\ref{sec:impl}.

The second contribution is empirical. I used the package to run a four-vendor, five-subject MMLU sweep: 50 questions per subject, 24 permutations per question, four vendors spanning a public capability gradient, total 24{,}000 API calls under temperature-0 generation. Predictions about which cells would show bias and which would not were locked into a pre-registration document whose SHA-256 hash was posted publicly before half the data was observed. The sweep shows that position bias is statistically detectable only within a roughly 60--95\% base-accuracy band, that the detectable cells separate into two mechanism types with distinct shape signatures, and that the public benchmark accuracy of current frontier-tier models places them above the detection band on every MMLU subject I ran. Results and pre-registration verification are in Section~\ref{sec:results}.

The third contribution is theoretical. A three-band capability model organises the findings. Low-tier models show processing-load dominance with a monotone signature on most subjects. Mid-tier models show content-ambiguity signatures on subject-intrinsic noisy domains while staying inactive elsewhere. High-tier models show no detectable bias because the benchmark cannot resolve them. The model is post-hoc relative to the pre-registration and is presented as such, with implications for benchmark design and vendor-claim interpretation. Section~\ref{sec:implications} discusses these, including a calibration-first scout protocol (\texttt{diagnose\_band}), a three-question filter for reading vendor no-bias claims, a lightweight pre-registration protocol that doesn't require a registered-report venue, and an analysis of how the wall-clock cost asymmetry across vendors interacts with the political economy of independent evaluation.

\subsection{Relation to prior work}

The methodological parent of this paper is arXiv 2606.26185, \emph{Necessary but Not Sufficient}~\citep{tamba2606}, which characterised the ceiling-effect side of position-bias detection: at high model capability, bias signals saturate and become uninformative as a measure of model character. The present work extends that result by reporting the floor side, where processing-load dominance swamps the subject-specific bias signal, and by characterising the middle band --- the Goldilocks zone --- where bias is genuinely measurable. The two papers together bracket what current measurement instruments can and cannot say about position bias as a function of model capability.

In the position-bias literature proper, \citet{zheng2024} reported that LLMs are not robust selectors across answer orderings. \citet{pezeshkpour2024} characterised sensitivity to order. \citet{wang2024beyond} surveyed the field and noted the role of position cues. All three rely primarily on single-shuffle or pairwise comparison protocols. I treat their findings as the empirical starting point that motivates an exhaustive-permutation instrument; the present work doesn't contest their position-bias claims but reports them with a finer-grained instrument.

The inspect\_ai framework~\citep{inspect_ai}, developed by the UK AI Safety Institute and external contributors, provides the substrate for the present work. Its dataset and solver abstractions let \texttt{inspect\_permute} operate as a strict drop-in replacement, and its eval-log format makes the per-run records that the present analysis depends on natively available.

I borrow the concept of \emph{item discriminability} from Item Response Theory~\citep{lord1952,birnbaum1968} as a conceptual frame for the Goldilocks zone, with the explicit caveat (Section~\ref{sec:design}) that the formal IRT apparatus does not transfer to language models. The structural parallel --- a measurement instrument failing to discriminate at either end of the latent-variable scale --- is well established in the psychometrics tradition, and I adopt the vocabulary because it makes the failure mode legible to readers from that tradition.

The cross-domain framing of measurement-instrument capture is treated extensively in Sismondo's \emph{Ghost-Managed Medicine}~\citep{sismondo2018}, which serves as the conceptual reference for Section~\ref{sec:implications}'s reading of cost asymmetry as an eval-ecosystem-level problem rather than a per-study one. I extend this reading to LLM evaluation in concurrent work~\citep{tamba_gmg}; the present paper restricts itself to the methodological and empirical contributions.

\section{The permutation diagnostic --- design}\label{sec:design}

\subsection{Why permutations rather than shuffles}

Most published position-bias measurements on MCQ benchmarks use a single shuffle. Each question gets presented to the model once, with the answer order drawn randomly~\citep{zheng2024,pezeshkpour2024,wang2024beyond}. The vendor reports an accuracy number and a permutation invariance number, usually as a single proportion. The protocol is cheap. It also confounds three sources of variation in the model's response.

The first is genuine position bias: the model systematically prefers some presented positions over others, even controlling for content. The second is question-level content noise: some questions are harder than others, and a single shuffle samples one ordering out of $k!$ without telling us whether the model's answer would have changed under a different one. The third is sampling noise on the eval set itself. With $N$ questions, the standard error on accuracy alone is already $\sqrt{p(1-p)/N}$, which dominates anything the position effect might add once $V$ falls below about 0.10.

A single-shuffle protocol cannot pull these three apart. If the model gets 80\% right under one shuffle and 78\% under another, the gap could be position bias on a few sensitive questions or noise on the eval set. Worse, the protocol cannot distinguish ``the model is biased'' from ``the benchmark exposes bias,'' which turns out empirically to be the distinction that matters (Section~\ref{sec:goldilocks}).

Exhaustive permutation is the natural alternative. For a $k$-option question I run all $k!$ orderings against the same model and record which presented position the model chose each time. The contingency table over (presented position of correct answer) $\times$ (correct / incorrect) is no longer a single draw from a noisy process. It is a complete enumeration of the response surface under the $k!$ permutations. For $k = 4$, the case for MMLU and most MCQ benchmarks, that means 24 calls per question. The cost is exactly $k!$ times a single-shuffle protocol, manageable for the question counts typical of bias studies.

\subsection{Epoch-as-permutation-slot}

The inspect\_ai framework already supports running each sample multiple times via its \texttt{epochs} parameter, originally designed for reducer-style accuracy estimation. I reuse the mechanism for a different purpose: each epoch carries a different permutation index. In exhaustive mode the index is \texttt{(epoch - 1) \% k!}, so the user requests \texttt{Epochs(k!)} and gets every permutation exactly once per question. In sample mode the index is drawn from a \texttt{(seed, sample\_id, epoch)} triplet via Python's \texttt{Random}, so each \texttt{(question, epoch)} pair maps to a deterministic permutation, but the permutation is not constrained to the exhaustive cycle.

This choice keeps \texttt{inspect\_permute} a strict drop-in for the framework's \texttt{multiple\_choice} solver. Datasets need no modification. The scoring pipeline is unchanged downstream of the new solver. The existing \texttt{original\_position} invariant on each \texttt{Choice} continues to hold, so the framework's built-in unshuffling logic still works on the eval log. The only downstream component that changes is the scorer, which I replace with one that records per-run position metadata. Everything else --- dataset loader, prompt template, model adapter, result writer --- is the same code as a non-permuted run.

The deterministic mapping has two practical advantages beyond architectural cleanliness. First, the permutation applied to question $q$ at epoch $e$ can be reconstructed from the eval log without storing the permutation explicitly. Second, the same \texttt{(seed, sample\_id, epoch)} triplet produces the same permutation across runs, which makes the diagnostic reproducible in the strict sense: an independent re-runner gets bit-exact agreement on which permutations were tried.

\subsection{Why Cram\'er's V}

The chi-squared test on the contingency table answers the question I actually care about: are accuracy and presented position independent? Under the null of independence, the test statistic has a known distribution, and the $p$-value gives a calibrated decision rule.

But $p$ alone is not enough to compare cells across vendors or subjects. The chi-squared statistic scales with $N$, so a marginal effect on a large eval set produces the same $p$-value as a large effect on a small one. For cross-cell comparison the report needs an effect size invariant to sample size.

Cram\'er's V provides exactly this. $V = \sqrt{\chi^2 / (N \cdot (\min(r,c) - 1))}$. For $k \times 2$ tables with $N = 1{,}200$ runs per cell, the formula reduces to $V = \sqrt{\chi^2 / N}$. $V$ ranges from 0 (perfect independence) to 1 (perfect association). It does not depend on the chi-squared distribution's degrees of freedom and so makes cells with different $k$ directly comparable.

I deliberately avoid the Yates continuity correction for tables with $\text{dof} \geq 2$, because the correction's conservative bias becomes substantial on the $k=4$ tables I use and would not match a literature comparison anyway. Tables that fall to the $2 \times 2$ case (degenerate cell) are reported with $V = 0$ rather than corrected, since the chi-squared statistic itself is then unreliable for inferential use.

\subsection{Shape diagnostics}

The chi-squared test reports whether bias is present; it does not report what kind. Position bias literature has noted at least two patterns~\citep{wang2024beyond,pezeshkpour2024}: a monotone preference for early presented positions (the model commits to the first plausible option) and a non-monotone anchoring effect (the model avoids the final option even when it has spent ample reasoning budget on the question). The contingency table contains the information needed to distinguish these. The chi-squared statistic compresses it away.

I add Spearman's rank correlation $\rho$ between presented position $(0, 1, \ldots, k-1)$ and accuracy at that position as a shape diagnostic. A model whose accuracy decreases monotonically with position has $\rho \to -1$. A model whose accuracy drops only at the final position has $\rho$ near 0 with possibly a small magnitude. A model with no positional structure has $\rho \approx 0$ and $V \approx 0$ together. The combination of $V$ and $\rho$ separates the two known patterns more reliably than either alone.

I use $\rho$ descriptively, not inferentially. With $k = 4$ positions, the Spearman test cannot reach conventional significance even for $\rho = -1.0$. The number of items is too small. The bootstrap CI on $\rho$ computed over the per-run records (1{,}000 resamples) is therefore an honest precision statement, not a hypothesis test. A cell with $V \geq 0.05$ and $p < 0.05$ is classified by the value of $\rho$ alone, with no claim that the $\rho$ value itself is statistically distinguishable from any other.

\subsection{An IRT parallel, used with care}\label{sec:irt}

The phenomenon I observe --- that bias signal magnitude depends on a measurement instrument's resolution relative to the underlying construct --- has a well-developed precedent in Item Response Theory. \citet{lord1952} and \citet{birnbaum1968} characterised item discriminability as the property of a test item that determines how much information the item provides about a latent ability. Items with low discriminability are uninformative regardless of the test-taker's ability. Items with high discriminability produce sharp differences in response probability across the ability range.

The structural parallel is direct. A bias-detection instrument with low discriminability --- for instance, an easy benchmark --- cannot tell a strongly biased model from an unbiased one, because both produce nearly identical surface behaviour. A high-discriminability instrument can. The Goldilocks zone I report in Section~\ref{sec:goldilocks} is in this sense an empirical statement about the discriminability of standard MMLU subjects as bias detectors.

I borrow the conceptual frame from IRT but not the formal model. Classical IRT presupposes a latent trait --- the test-taker's ability --- that gets estimated through the response pattern. Language models are not test-takers in this sense. They have no stable latent ability the response pattern reveals, only a parameterised function whose outputs are conditional on the prompt. The mechanism by which position cues affect the response is also different. In human testing, latent-trait estimation noise produces a response distribution that the bias acts on probabilistically. In LLMs, the position information enters through attention weights on the prompt tokens and influences generation directly. The phenomena are parallel. The mechanisms are independent. I adopt the IRT vocabulary to make the underlying problem legible to readers from the measurement tradition, but the inferential apparatus of IRT does not transfer.

\section{Implementation}\label{sec:impl}

\subsection{Package overview}

The diagnostic is implemented as \texttt{inspect\_permute}, an open-source Python package that operates as a drop-in extension to the inspect\_ai evaluation framework~\citep{inspect_ai}. The package exposes four objects that compose with inspect\_ai's existing \texttt{Task}, \texttt{Solver}, and \texttt{Scorer} abstractions, plus a planned diagnostic helper.

The solver, \texttt{permute\_choices}, replaces the standard \texttt{multiple\_choice} solver. It reads inspect\_ai's per-sample \texttt{epoch} counter and applies a deterministic permutation of answer order tied to the \texttt{(sample\_id, epoch)} pair. In exhaustive mode it cycles through all $k!$ orderings of a $k$-option question as the epoch index increments. In sample mode it draws a random permutation seeded by the pair, allowing reproducible partial coverage when full enumeration is too expensive. The solver preserves inspect\_ai's \texttt{original\_position} invariant on each \texttt{Choice}, so the framework's standard \texttt{unshuffle\_choices} continues to work and the eval log records the permutation as part of \texttt{state.metadata}. This last property is what makes the rest of the pipeline reconstructable from logs alone.

The scorer, \texttt{position\_bias\_score}, evaluates correctness against the un-shuffled target and writes per-run metadata: the presented position of the correct answer, the presented position the model picked, the permutation vector applied, the epoch index, and the question's number of options. These five fields are the per-run records the report module aggregates.

The report module, \texttt{generate\_bias\_report}, consumes the eval logs and produces the cell-level statistics: the contingency table over (correct\_position, correct/incorrect), the Pearson chi-squared statistic with the matching $p$-value, the Cram\'er's V effect size, and the per-position accuracy and pick-distribution breakdowns. The report function is total. It returns the same dict whether the table is degenerate (perfect model, perfect-failure, or empty rows), with the degenerate cases returning $\chi^2 = 0$, $p = 1$, and $V = 0$ rather than raising.

The shape diagnostics module, \texttt{shape\_diagnostics}, adds Spearman $\rho$ between presented position and accuracy at that position, bootstrap 95\% confidence intervals on both $V$ and $\rho$, and the mechanism classifier described in Section~\ref{sec:classifier}. It accepts either raw eval logs (for full bootstrap) or a pre-computed matrix (for fast post-hoc classification across a sweep).

A planned \texttt{diagnose\_band} API (v0.2 design, not used in the present sweep) runs a short scout: two orthogonal permutations applied to 25 randomly selected questions. The estimate of base accuracy does not get contaminated by the position bias the full sweep is supposed to detect. The two permutations are deliberately the identity and the reverse, so any first-order position effect on the scout cancels across the pair. The resulting band classification (high, middle, low) lets users skip an expensive full sweep on cells the tool cannot resolve. The cost pattern I encountered live during the present work (Section~\ref{sec:cost}) is the motivating use case.

\subsection{Reproducibility infrastructure}

The package was designed for reproducibility before correctness. Each tool decision was chosen to make an independent re-run easy to verify against the original. Generation temperature is set to 0 for vendors that support it. Concurrency is fixed at 10 per cell. Transient errors are retried up to two times, a setting that proved consequential when Anthropic credit ran out mid-sweep (Section~\ref{sec:robustness}) but does not affect the bias measurement on a clean run.

Bootstrap confidence intervals on both $V$ and $\rho$ use 1{,}000 resamples of the per-run records with replacement, with a fixed seed for the resample sequence. Inspect\_ai's \texttt{.eval} log files (one per cell) are preserved alongside the bias report so any cell can be re-analysed without re-running the API calls.

The same protocol that supports replicability also supports pre-registration. Before the sweep began I drafted a set of falsifier predictions in \texttt{preregistration.md}, including numerical decision rules and per-cell expected $V$ ranges (Appendix~\ref{app:prereg} reproduces both). The file's SHA-256 hash was committed to a local git tag (\texttt{preregistration-amended-2026-06-30}) and posted to LinkedIn at 2026-06-30 16:28 JST. Two of the four vendors and three of the five subjects had no data at the moment the hash was published. The git tag preserves the file at the hashing moment. The LinkedIn post timestamp is the public anchor. The matrix.json snapshot at that commit shows only the 10 cells that had already been run. A third-party reviewer can verify the chain in three commands (Appendix~\ref{app:prereg}).

I treat this combination --- local git tag, public hash post, on-disk matrix snapshot at hash time --- as a lightweight pre-registration protocol that any researcher can replicate without paywall access to a registered-report venue. The protocol is incidentally a contribution of the present work and is discussed in Section~\ref{sec:implications}.

\subsection{A classifier flaw discovered through the sweep}\label{sec:classifier}

The v0.1 mechanism classifier was defined ahead of data collection. It used a single effect-size floor and a single Spearman cut to assign each cell to one of three labels. Inactive if $V < 0.05$. Processing\_load if $V \geq 0.05$ and $\rho \leq -0.5$. Content\_ambiguity if $V \geq 0.05$ and $\rho > -0.5$. The floor at $V = 0.05$ was chosen as the rule-of-thumb point below which the chi-squared signal could plausibly be attributed to sampling noise on tables of this size.

A single cell exposed the rule's blind spot. The grok-3 cell on elementary\_mathematics returned $V = 0.058$, just above the floor, with Spearman $\rho = -0.32$, just above the processing\_load cut. The v0.1 classifier therefore labels the cell \texttt{content\_ambiguity}. But the cell's $p$-value is 0.26, and its picked-when-wrong distribution is 2/2/2/2 across the four positions --- perfectly uniform on the eight runs the model got wrong out of 1{,}200. There is no statistical evidence of position bias at this cell. The classification is a parsing artefact of the rule.

The fix is to add an explicit gate on the chi-squared test itself. If $p \geq \alpha$ or $V < V_{\text{floor}}$, classify as inactive regardless of $\rho$. The refined classifier is what \texttt{shape\_diagnostics.classify} implements (with default $\alpha = 0.05$, $V_{\text{floor}} = 0.05$, $\rho_{\text{processing}} = -0.5$). The preregistered $V$-only rule is preserved verbatim in the package as \texttt{classify\_preregistered} for reproducibility of the v0.1 verdict.

I disclose the flaw and the fix here rather than silently substituting the refined classifier, for three reasons. First, the per-cell prediction hit rate reported in Section~\ref{sec:falsifier} is computed against the preregistered rule, not the refined one --- including the grok-3 elementary cell as a technical miss --- so the hit rate is honestly degraded by the boundary case. Second, the discovery is an example of dogfooding: the rule's blind spot was exposed by the sweep it was meant to evaluate, and the resulting refinement is a deliverable for downstream users of the package. Third, the refinement is itself a small methodological contribution that I want readers to identify, evaluate, and override if better thresholds become apparent. The package documentation flags both classifiers and explains when each applies.

\section{Results}\label{sec:results}

\subsection{Sweep design}

The sweep applied the permutation diagnostic across four vendors and five MMLU subjects, with 50 questions per subject and exhaustive 24-permutation coverage per question. Total: $4 \times 5 \times 50 \times 24 = 24{,}000$ API calls. Vendors spanned a public capability range at the time of the run: \texttt{openai/gpt-4o-mini}, \texttt{anthropic/claude-haiku-4-5}, \texttt{google/gemini-2.5-flash}, \texttt{xai/grok-3}. Subjects spanned a difficulty gradient and included both symbolic-reasoning and humanities domains, listed easiest to hardest by typical model accuracy: elementary\_mathematics, college\_mathematics, formal\_logic, professional\_medicine, philosophy.

Generation temperature was set to 0 wherever the vendor API supported it, so the response variance is attributable to permutation order rather than sampling stochasticity. The solver applied each of the 24 permutations exactly once per question in exhaustive mode, with permutation index deterministically tied to the (sample\_id, epoch) pair. The run used \texttt{max\_connections=10} and \texttt{retry\_on\_error=2}. The latter became consequential during the Anthropic credit interruption discussed in Section~\ref{sec:robustness}.

Falsifier predictions and per-cell expected $V$ ranges were locked into a preregistration document. Its SHA-256 hash was published on LinkedIn at 2026-06-30 16:28 JST, before the first Google or xAI cell completed. The corresponding git tag (\texttt{preregistration-amended-2026-06-30}, commit \texttt{ad33083}) records the file state at that moment. The verification procedure is in Appendix~\ref{app:prereg}.

\subsection{Headline matrix}\label{sec:headline}

\begin{figure}[H]
\centering
\includegraphics[width=0.95\textwidth]{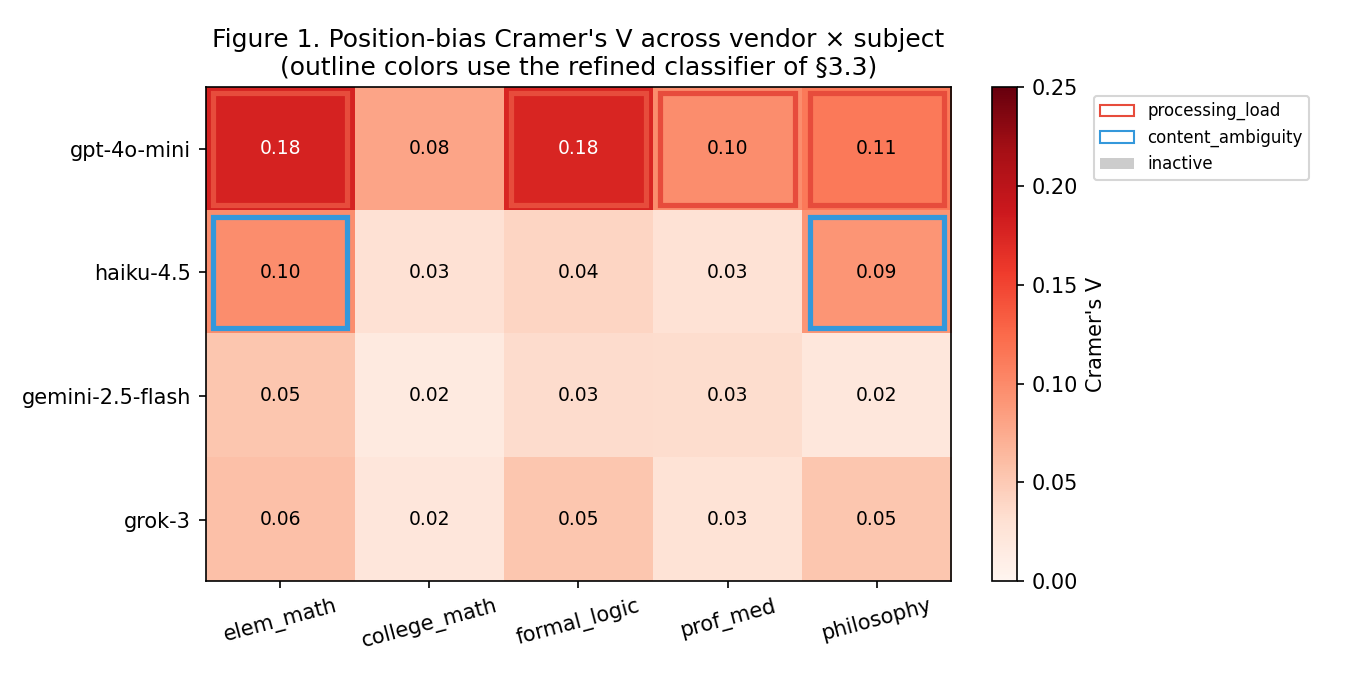}
\caption{Position-bias Cram\'er's V across vendor $\times$ subject. Outline colors use the refined classifier of Section~\ref{sec:classifier}: red for \texttt{processing\_load}, blue for \texttt{content\_ambiguity}, no outline for \texttt{inactive}.}
\label{fig:heatmap}
\end{figure}

The full vendor-by-subject matrix of effect sizes is in Figure~\ref{fig:heatmap}. Cells outlined in red are classified \texttt{processing\_load} by the refined classifier. Cells outlined in blue are \texttt{content\_ambiguity}. Cells without an outline are \texttt{inactive} (either $V < 0.05$ or $p \geq 0.05$ after bootstrap; see Section~\ref{sec:vrho} for the classifier and Section~\ref{sec:classifier} for its derivation).

\begin{itemize}
\item \textbf{gpt-4o-mini} registers significant position bias on four of five subjects, $V$ from 0.10 (professional\_medicine) to 0.18 (elementary\_mathematics, formal\_logic). The exception is college\_mathematics, where $V = 0.08$ falls just outside significance at $p = 0.057$. Every active gpt-4o-mini cell is \texttt{processing\_load}.
\item \textbf{claude-haiku-4-5} registers significant bias on two of five subjects: elementary\_mathematics ($V = 0.10$) and philosophy ($V = 0.09$). Both are \texttt{content\_ambiguity}. The remaining three subjects show $V \leq 0.05$ and $p > 0.4$.
\item \textbf{gemini-2.5-flash} shows $V \leq 0.05$ on every subject, with all $p$-values above 0.45. No bias detected.
\item \textbf{grok-3} shows the same pattern as gemini-2.5-flash, with $V \leq 0.06$ on every subject. Its elementary\_mathematics cell ($V = 0.058$, $p = 0.26$) sits closest to the floor and is examined further in Section~\ref{sec:vrho} as the boundary case that motivated the classifier refinement.
\end{itemize}

The matrix invites two competing readings. A vendor-tier reading: gpt-4o-mini has serious position bias, haiku has mild residual bias on subject-intrinsic edge cases, and the two flagship-tier models have no detectable bias. A capability-band reading: bias is visible only where the model's base accuracy leaves enough variance to detect it, and the flagship-tier cells are all base accuracy $\geq 95\%$. The remaining subsections build out the second reading.

All $V$ values, mechanism labels, and figure colourings reported in Sections~\ref{sec:headline}--\ref{sec:goldilocks} use the refined classifier of Section~\ref{sec:classifier} (with explicit $p$-value gating). The pre-registered falsifier predictions in Section~\ref{sec:falsifier} are scored against the original v0.1 classifier (V-floor only) so the verification metric is not retroactively improved by the refinement.

\subsection{Mechanism signatures}

\begin{figure}[H]
\centering
\includegraphics[width=0.95\textwidth]{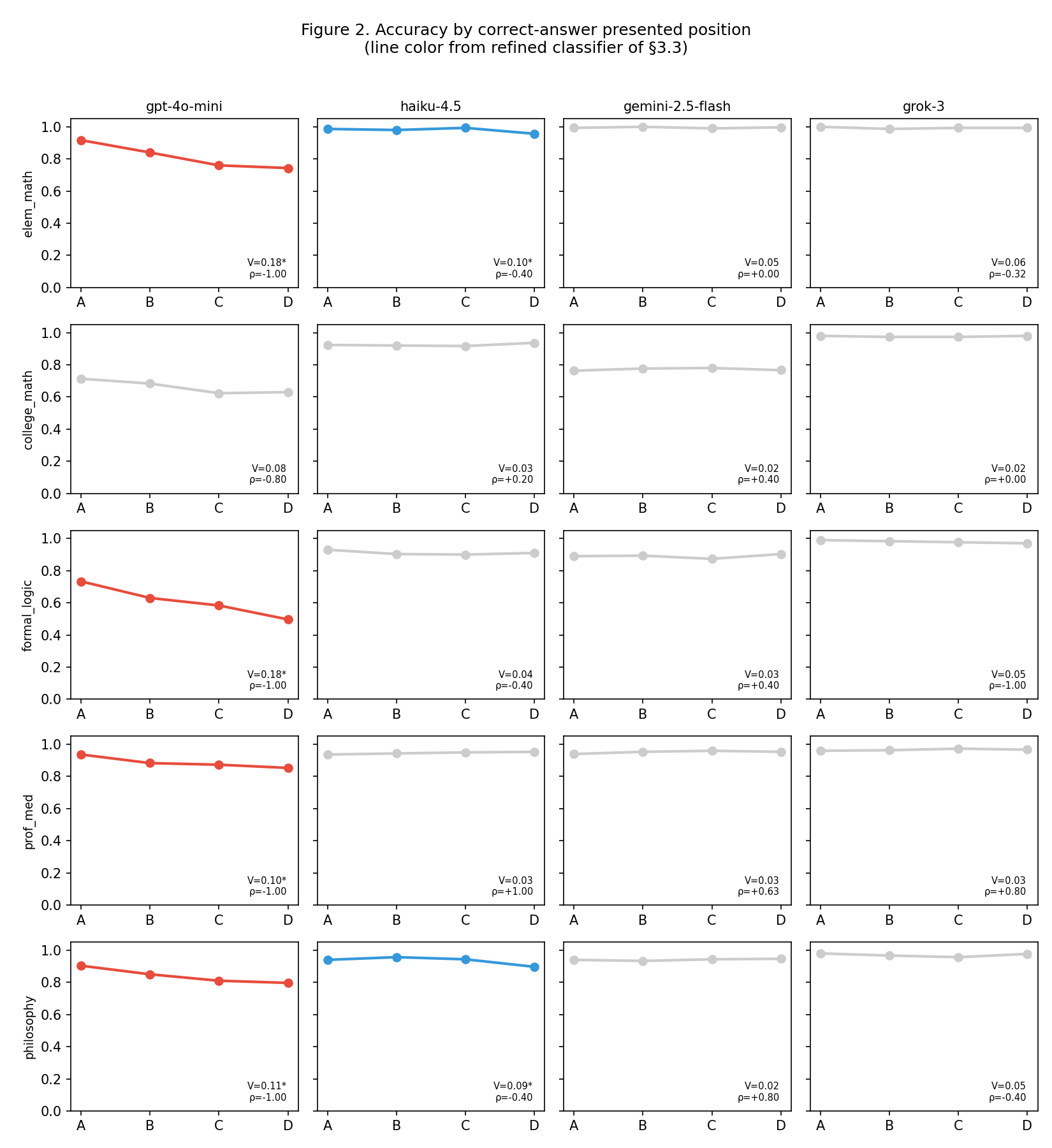}
\caption{Accuracy by correct-answer presented position. Line color from the refined classifier of Section~\ref{sec:classifier}. Red lines indicate \texttt{processing\_load} (monotone A\toarrow{}D decrease). Blue lines indicate \texttt{content\_ambiguity} (non-monotone D-drop). Grey lines indicate \texttt{inactive}.}
\label{fig:smallmultiples}
\end{figure}

Within the cells where bias is detected, the shape of the bias differs systematically by vendor. Figure~\ref{fig:smallmultiples} plots accuracy by presented position for all 20 cells.

In every gpt-4o-mini cell with significant bias, accuracy decreases monotonically from position A to position D. The clearest example is formal\_logic, where accuracy falls from 0.73 at A to 0.50 at D --- a 23-point gap inside a single question for the same content under different orderings. This monotone profile (Spearman $\rho = -1.0$ in all four active cells) is the signature for \texttt{processing\_load}: the model preferentially commits to the first encountered option when reasoning budget is exhausted, leaving later positions systematically underchosen even when correct.

The two active haiku cells show a different shape. Accuracy stays near ceiling at positions A, B, and C (range 0.94--0.99) and drops by 3 to 6 points at position D specifically. Spearman $\rho$ values are $-0.40$ in both cells, far from the $-1.0$ that processing\_load would produce. This is the \texttt{content\_ambiguity} pattern: when several options are individually plausible, a subject-intrinsic anchoring effect pulls the model away from the final option, even though the model has plenty of computational headroom on the rest of the question. The fact that this signature reproduces on two subjects within a single vendor (elementary\_mathematics and philosophy) suggests it is a real mechanism and not noise on a single cell.

The remaining ten cells, across gemini-2.5-flash and grok-3, show near-flat accuracy profiles. The largest within-cell range is 4 points (gemini $\times$ elementary, 0.987 at B to 0.997 at A). Most are below 2 points. Spearman $\rho$ values are dispersed without a consistent sign, ranging from $+1.0$ to $-1.0$ but constrained to a $|V| < 0.05$ regime where the underlying contingency table has too little variance for the rank correlation to be meaningful.

\subsection{Effect-size and shape coordinates}\label{sec:vrho}

\begin{figure}[H]
\centering
\includegraphics[width=0.85\textwidth]{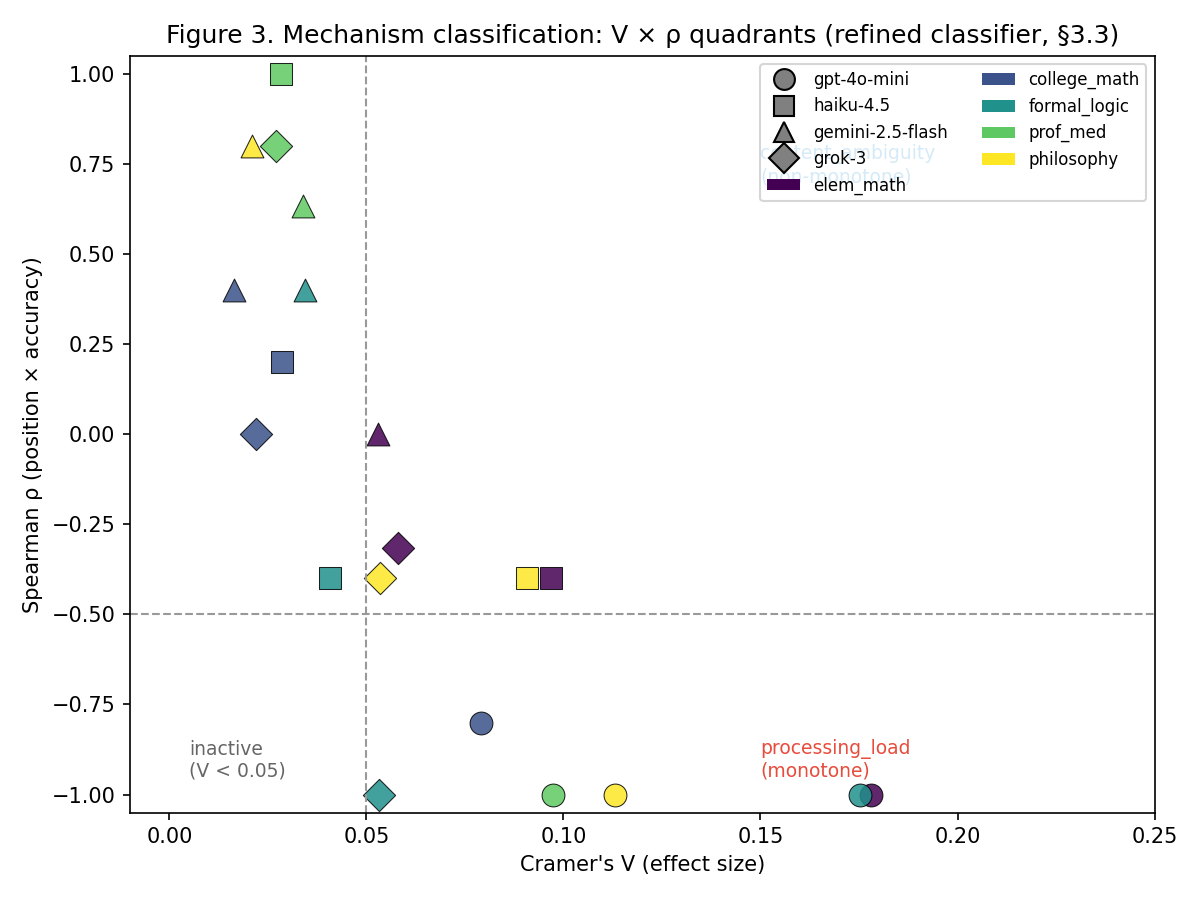}
\caption{Mechanism classification in the $V \times \rho$ plane. Vendor marker shapes: circle (gpt-4o-mini), square (haiku-4.5), triangle (gemini-2.5-flash), diamond (grok-3). Subject color from viridis. Dashed lines mark refined-classifier (Section~\ref{sec:classifier}) cutoffs at $V = 0.05$ and $\rho = -0.5$.}
\label{fig:vrho}
\end{figure}

Figure~\ref{fig:vrho} plots each cell in the $V \times \rho$ plane to bring the two-mechanism distinction into a single view. Vendor identity is encoded by marker shape; subject by colour.

The plane separates into three regions corresponding to the refined classifier (Section~\ref{sec:classifier}):

\begin{itemize}
\item \textbf{Bottom-right} ($V \geq 0.05$, $\rho \leq -0.5$): \texttt{processing\_load}, populated entirely by gpt-4o-mini cells.
\item \textbf{Right-middle} ($V \geq 0.05$, $-0.5 < \rho$): \texttt{content\_ambiguity}, populated by the two active haiku cells.
\item \textbf{Left} ($V < 0.05$): \texttt{inactive}, populated by every gemini and grok cell as well as the inactive haiku and gpt-4o-mini cells.
\end{itemize}

The plane also surfaces a boundary case that drove a methodological refinement. The grok-3 $\times$ elementary cell sits at $V = 0.058$ with $\rho = -0.32$, just inside the right-half by $V$ but well above the processing-load threshold by $\rho$. Its $p$-value is 0.26: no statistical evidence of bias at conventional significance. The v0.1 classifier (preregistered using only the $V \geq 0.05$ floor and the $\rho \leq -0.5$ cut) would label this cell \texttt{content\_ambiguity} purely on marginal $V$ and weak $\rho$. The refined classifier adds an explicit gate, $p \geq \alpha \implies \text{inactive}$, which correctly relegates the cell to the inactive region. Section~\ref{sec:classifier} documents this refinement as a contribution discovered through the sweep rather than imposed before it. The preregistered classifier's predictions are scored without modification in Section~\ref{sec:falsifier}.

The thresholds in the refined classifier --- $V_{\text{floor}} = 0.05$, $\rho_{\text{processing}} = -0.5$, $\alpha = 0.05$ --- are conventional rather than derived. Their choice is therefore a candidate confound. Appendix~\ref{app:sensitivity} reports a sensitivity sweep over $V_{\text{floor}} \in \{0.04, 0.05, 0.06\}$ and $\rho_{\text{processing}} \in \{-0.4, -0.5, -0.6\}$. Only the grok-3 elementary cell changes label under any perturbation, and only between \texttt{inactive} and \texttt{content\_ambiguity}. The matrix structure of Section~\ref{sec:headline} is invariant to these choices.

\subsection{The Goldilocks zone}\label{sec:goldilocks}

\begin{figure}[H]
\centering
\includegraphics[width=0.85\textwidth]{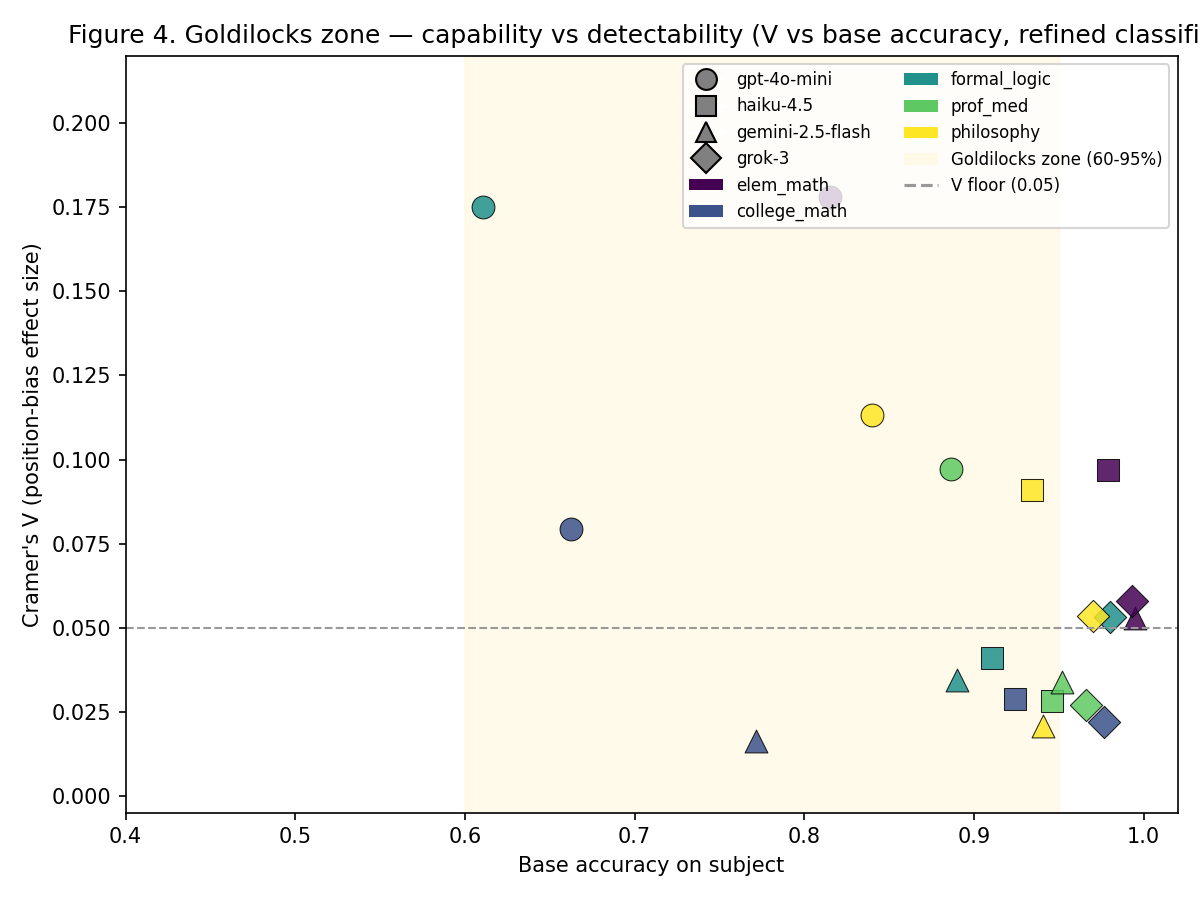}
\caption{Goldilocks zone --- capability vs. detectability. $V$ vs. base accuracy under the refined classifier. Shaded band marks the 60--95\% accuracy range within which position bias is statistically detectable.}
\label{fig:goldilocks}
\end{figure}

Both the vendor and the mechanism readings of the matrix collapse into a single observation when each cell is plotted against its base accuracy (Figure~\ref{fig:goldilocks}). The shaded region marks the band $0.60 \leq \text{accuracy} \leq 0.95$.

Every cell with detected bias lies inside the shaded band. Outside the band, on the high-accuracy side, twelve cells cluster below the $V = 0.05$ floor: nothing detectable. Inside the band, ten cells span the full range from $V = 0.02$ to $V = 0.18$, with the relationship between accuracy and bias intensity being non-monotonic --- gpt-4o-mini $\times$ formal\_logic at accuracy 0.61 and gpt-4o-mini $\times$ elementary\_math at accuracy 0.81 both reach $V = 0.18$.

The observation summarises as a Goldilocks zone for position-bias detection: position bias is statistically visible only when the model's accuracy on the benchmark leaves enough variance for the chi-squared test to find structure. Above the zone, ceiling effects compress the variance below the test's resolution. Below the zone (not represented in this sweep but predicted by extrapolation), the bias signal would be expected to give way to general capability noise.

The Goldilocks framing and the three-band capability model that motivates Section~\ref{sec:implications} emerged from post-hoc inspection of the matrix once all 20 cells were observed. The boundary case discussed in Sections~\ref{sec:classifier} and~\ref{sec:vrho} was the proximate trigger. The systematic direction of the pre-registered prediction misses (Section~\ref{sec:falsifier}) was the corroborating signal. None of the predictions in the pre-registered document anticipated the three-band structure, and Section~\ref{sec:falsifier} scores the predictions without it.

The implication is not that flagship-tier models lack position bias. It is that standard MMLU subjects are the wrong instrument to detect bias in flagship-tier models. The right reading of the gemini-2.5-flash and grok-3 rows in Figure~\ref{fig:heatmap} is ``no signal,'' not ``no bias.'' Answering the latter question would require benchmarks engineered to put flagship-tier models inside the zone, which Section~\ref{sec:implications} develops as a recommendation.

Two further points the figure makes visible. First, the upper edge of the zone is not strictly at accuracy 0.95: haiku $\times$ elementary at accuracy 0.99 and haiku $\times$ philosophy at accuracy 0.94 both register $V \approx 0.09$, suggesting that subjects with intrinsic decision noise can keep content\_ambiguity detectable even at near-ceiling accuracies. Second, the leftmost cells in the figure are at accuracy 0.61 (gpt-4o-mini $\times$ formal\_logic) and 0.66 (gpt-4o-mini $\times$ college\_math). The sweep did not include any model-subject pair below 0.60, so the floor edge of the zone is predicted but not directly observed.

\subsection{Pre-registered falsifier verification}\label{sec:falsifier}

Three falsifiers were locked into the preregistration document with explicit numerical decision rules, allowing each falsifier to be scored as confirmed, partially failed, rejected, or inconclusive without post-hoc adjudication. The outcomes are summarised here; the decision rules are reproduced verbatim in Appendix~\ref{app:prereg}.

\paragraph{Falsifier 1 --- college\_math null across all vendors.} Predicted: college\_math behaves as a decision-determinate domain and produces $V < 0.10$ and $p > 0.05$ for all four vendors. Observed: $V$ values 0.08, 0.03, 0.02, 0.02 with $p$ values 0.057, 0.83, 0.80, 0.80. All four vendors satisfy the null condition. \textbf{Verdict: CONFIRMED.}

\paragraph{Falsifier 2 --- philosophy non-monotone for flagships.} Predicted: if bias is detected on philosophy in either gemini-2.5-flash or grok-3, the bias would have $\rho > -0.5$ (non-monotone, content\_ambiguity signature). Observed: neither flagship vendor showed detected bias on philosophy ($V = 0.02$ and $V = 0.05$ respectively, both $p > 0.10$), leaving the falsifier silent on shape. \textbf{Verdict: INCONCLUSIVE.} The inconclusive verdict is consistent with the 3-band model rather than evidence against it; the model predicts no bias at high tier regardless of mechanism, so the absence of a shape signal is expected.

\paragraph{Falsifier 3 --- flagship ceiling on elementary\_mathematics.} Predicted: gemini-2.5-flash and grok-3 would both show $V < 0.10$ on the easiest subject, signalling that they had saturated the benchmark. Observed: $V = 0.05$ for gemini, $V = 0.06$ for grok. Both fall comfortably below the threshold. \textbf{Verdict: CONFIRMED.}

Two confirmed falsifiers and one inconclusive falsifier together leave the underlying framework intact while declining to grant Falsifier 2 the strongest form of evidence it could have provided.

Per-cell prediction outcomes for the ten unobserved cells (gemini $\times$ 5 subjects, grok $\times$ 5 subjects) are reported as follows. The per-cell predictions specified both an expected $V$ range and an expected mechanism class. Of the ten cells, the $V$ range prediction is correct for six and the mechanism class prediction is correct for five. The five missed mechanism predictions all share the same structure: the preregistration predicted content\_ambiguity active on philosophy for both flagship vendors (and weak processing\_load on formal\_logic for grok), but the observed cells fell to inactive in each case. The systematic direction of the misses --- predicted active, observed inactive at high tier --- is the empirical signal that motivated the 3-band model in Section~\ref{sec:goldilocks}. At the time the predictions were written, I had not appreciated that the content\_ambiguity mechanism would also be capability-gated. The misses are reported here unmodified. The post-hoc framework adjustment is disclosed in Section~\ref{sec:goldilocks} rather than absorbed into the verification metric.

\subsection{Robustness --- credit exhaustion did not contaminate measurement}\label{sec:robustness}

During the first sweep, the Anthropic API credit balance ran out partway through the anthropic/professional\_medicine cell. With \texttt{retry\_on\_error=2} enabled, every blocked sample kept retrying until I noticed the low-balance email and topped the account up. The five anthropic cells produced complete output ($n = 1{,}200$ runs each), but the wall-clock timings were entangled with the retry backoff window.

To verify the measurement itself had not been distorted, I backed up the contaminated v1 results, deleted the cached \texttt{bias\_report.json} files, and re-ran all five anthropic cells against a credit-positive account. The maximum $|V_{v2} - V_{v1}|$ across the five cells was 0.0058, observed on formal\_logic. On philosophy the two runs agree to four decimal places ($V = 0.0909$ in both). Per-position accuracy values agree within 0.003. The retry mechanism preserved measurement integrity completely under the credit interruption. The only quantity affected was wall-clock timing, which I set aside for the cost analysis below.

This is a small finding but worth recording. The extension does not require uninterrupted billing to produce trustworthy measurements, as long as the retry mechanism is enabled.

\subsection{Wall-clock cost asymmetry (exploratory, not pre-registered)}\label{sec:cost}

The same protocol carries strikingly different wall-clock cost across vendors. Per-cell wall-clock times averaged over the five subjects, using clean runs for all vendors:

\begin{itemize}
\item openai/gpt-4o-mini: about 2.0 minutes per cell.
\item anthropic/claude-haiku-4-5: about 4.3 minutes per cell.
\item google/gemini-2.5-flash: about 15 minutes per cell (range 5--37 minutes; the upper figure is college\_math).
\item xai/grok-3: about 5 minutes per cell (varies with reasoning trace length).
\end{itemize}

The disparity tracks the proportion of reasoning (``thinking'') tokens reported by each vendor's usage breakdown. For gemini-2.5-flash on college\_mathematics, 85\% of total tokens were reasoning tokens, against 0\% for both openai/gpt-4o-mini and anthropic/claude-haiku-4-5. The peak wall-clock asymmetry is $18.5\times$ (gemini college\_math vs openai college\_math, both 1{,}200 calls); the average is $7.5\times$. Inspection of the sweep log for the gemini section confirmed that the slowdown is driven entirely by reasoning-trace generation rather than rate-limit retries or error backoff.

This is a methodological point with practical bite. Independent replication of a position-bias sweep on a reasoning-trace model demands an order of magnitude more compute time and proportionally more cost than on a non-reasoning model. The barrier to replication is not knowledge of the protocol or access to the API. It is the eval-time compute footprint, which inherits the asymmetry of the underlying inference platforms. Resource-rich evaluators can sweep all four vendors comfortably. An independent evaluator with a single workstation will find the gemini sweep alone consuming the majority of the run time. Section~\ref{sec:implications} returns to this as a contributing factor to evaluation ecosystem capture.

\section{Implications}\label{sec:implications}

\subsection{Calibrating a benchmark to a model tier}

The most actionable consequence of the Goldilocks zone observation is that an evaluation designer cannot answer ``does model $M$ have position bias on benchmark $B$'' without first asking ``is $M$'s base accuracy on $B$ inside the detection window.'' The matrix in Section~\ref{sec:results} makes the point concretely. gemini-2.5-flash and grok-3 sit above 0.95 on every MMLU subject I ran, and at that accuracy level no permutation diagnostic can produce a verdict beyond ``not measurable on this instrument.'' Reading the absence of a chi-squared signal as evidence the model is unbiased is a category error.

The remedy is a calibration step before the full sweep. I propose \texttt{diagnose\_band}, a short scout protocol that estimates a model's base accuracy on a benchmark using a small number of API calls and classifies the resulting cell into one of three bands. High (accuracy $\geq$ 0.95, ceiling-saturated, full sweep skipped). Middle (0.60--0.95, full sweep recommended, results interpretable). Low (below 0.60, processing-load dominant, content-ambiguity signal not separable). The scout uses two orthogonal permutations applied to 25 questions: the identity permutation and the reverse. Any first-order position effect cancels across the pair, so the accuracy estimate is not contaminated by the very bias the full sweep is supposed to detect. The two-permutation scout costs 50 calls, about 1\% of a full sweep at standard $k=4$.

The scout produces a recommendation rather than a measurement. A high-band classification turns the immediate research question --- does this model have position bias on this benchmark --- into a different question: which benchmark would put this model inside the zone. The diagnostic does not answer that question by itself, but it lets the designer stop spending 30-minute Gemini calls on cells that will return no usable signal. The cost asymmetry result of Section~\ref{sec:cost} makes the saving practically important. A single high-band cell that runs to completion on a thinking-trace vendor can consume more wall-clock than five low-band cells on a non-reasoning vendor.

\subsection{A flowchart for reading vendor bias claims}

The matrix structure also gives evaluators a sharper way to read vendor marketing. A claim of the form ``our model has accuracy $X$ and no position bias on benchmark $B$'' should be parsed as three questions, in order.

First, is the model's accuracy at $X$ above 0.95? If so, the no-bias claim is unverifiable on $B$. The vendor has measured saturation, not bias.

Second, if the accuracy sits in the 0.60--0.95 band, was the bias measurement done with permutation enumeration or with a single shuffle? A single-shuffle no-bias claim is at best a non-significance from one draw of $k!$ orderings, not a verified null.

Third, if the measurement was done correctly and the accuracy is inside the band, what was the effect size and the shape? Cram\'er's V near 0.05 with $\rho$ near 0 is consistent with the null. Cram\'er's V near 0.05 with $\rho$ near $-1.0$ is suggestive of weak processing-load and should not be dismissed.

This is not a flowchart that demolishes vendor claims. Many will pass it. It is a flowchart that separates verifiable claims from unverifiable ones, and the second category is more common than the first under current measurement practice.

\subsection{The cost asymmetry is an eval-ecosystem problem}

Section~\ref{sec:cost} reported that the same protocol costs $7.5\times$ more wall-clock on gemini-2.5-flash than on gpt-4o-mini on average, with a peak of $18.5\times$ on the same college\_math cell. The cost is not paid by the vendor. It is paid by anyone who wants to evaluate the vendor. An independent researcher running a four-vendor sweep on a single workstation will find the gemini cells alone consuming most of the run time.

This matters for the ecology of LLM evaluation. Pre-deployment safety evaluation, as currently practised, is performed primarily by the model vendors themselves and by national AI safety institutes with privileged access to internal model variants. The set of independent third parties able to replicate or contest these evaluations is constrained by what they can afford in compute time. As models shift toward extended reasoning traces, the asymmetry compounds. Each replication step becomes more expensive without the underlying methodology becoming any harder. The result is an evaluation ecosystem in which the vendors set the cost basis for the very tests they are being evaluated against.

I do not propose a remedy here. The point is that the cost asymmetry is not a side observation of the present sweep. It is a structural feature of the eval landscape that any future bias-detection tooling will inherit, and the permutation diagnostic is one of the cheaper such tools per result.

\subsection{Preregistration without a registered-report venue}

The pre-registration protocol I used here --- local git commit with file-level SHA-256, public hash post to a dated platform, an on-disk data snapshot at hash time --- is portable. It needs no relationship with a registered-report journal, no editorial review of the prediction, no platform fee. The git tag is the file-state anchor. The public hash post is the timestamp anchor. The matrix snapshot at that commit is the observation-state anchor. Any reader can verify the three in sequence using \texttt{git checkout}, \texttt{sha256sum}, and a JSON read of the matrix file. The verification procedure is reproduced in Appendix~\ref{app:prereg} and takes under a minute on a modern laptop.

I used LinkedIn for the public hash post because I already had an active account with timestamped posts, but the protocol does not depend on the platform. Any service that produces a public, timestamped, non-revisable artifact would serve. The point is that pre-registration is a habit of disclosure, not a feature of a venue. The cost of the protocol is the discipline to commit predictions before observation, which is the actual scarce resource.

The protocol also accommodates partial pre-registration. In the present work, the falsifier rules and the per-cell expected ranges were committed before observation, but the mechanism classifier itself was refined post-hoc (Section~\ref{sec:classifier}). The git history records which classifier version was used to score which prediction, so the methodological boundary between pre-registered and exploratory work is visible in the commit log. No reader has to take the author's word for the boundary. The verification procedure surfaces it directly.

\subsection{What we did not measure}

A first limitation worth naming directly. The sweep used five MMLU subjects, a single benchmark family. Whether the Goldilocks zone width and the band thresholds I observe transfer to non-MMLU benchmarks (BBH, AGIEval, FrontierMath, TruthfulQA) is an empirical question, not an answered one. Standard MMLU is saturated for current frontier-tier models on most subjects, which is precisely the condition that produces the high-band ceiling I report. Benchmarks with denser hard items would put more cells inside the zone and yield more independent observations of the two mechanisms. The framework as stated should be tested on at least one non-MMLU corpus before any general claim about benchmark calibration is taken to apply.

Beyond benchmark scope, three open questions are exposed by the present sweep and left for future work.

The upper edge of the Goldilocks zone is subject-dependent. haiku $\times$ elementary at accuracy 0.99 and haiku $\times$ philosophy at accuracy 0.94 both register content-ambiguity bias, suggesting the edge depends on intrinsic decision noise in the subject domain rather than on a fixed accuracy threshold. Characterising this dependence requires a denser sampling of (subject, accuracy) combinations than five MMLU subjects can support.

The floor edge of the zone is predicted by extrapolation but not directly observed. The lowest base accuracy in the sweep is 0.61 (gpt-4o-mini on formal\_logic). A model-benchmark combination below 0.60 should, on the framework, show neither processing\_load nor content\_ambiguity in a separable form, but I have no empirical point that anchors the prediction.

The reasoning-trace vendors (gemini-2.5-flash, grok-3) showed no detectable bias on any subject in the sweep. Whether the absence reflects genuine bias mitigation, ceiling saturation, or bias laundering --- where the thinking trace contains the position-biased reasoning but the surface output overrides it --- cannot be settled without access to the trace itself. A version of \texttt{position\_bias\_score} that records reasoning-trace metadata, where the vendor exposes it, is a natural v0.2 addition.

These three questions together motivate a v1.0 study against frontier-tier models on benchmarks that bring them into the Goldilocks zone. Realising that study requires access to flagship-tier model APIs and to evaluation benchmarks at the right capability level. The access question is a coordination problem rather than future work in the strict sense: the diagnostic exists, the protocol is reproducible, and the gap is now a question of which evaluator gets to run it on which model.

\subsection{Two papers, one bracket}

The present work closes a side of the detection window that an earlier paper opened. \citet{tamba2606} reported that position-bias signals saturate at high model capability and become uninformative as a measure of model character --- the ceiling-effect side of the window. The present work reports a parallel saturation at low model capability, where processing-load swamps any subject-specific bias signal --- the floor-effect side. Between the two, the detection window is narrower than the literature has assumed, and the location of the window is itself a function of the (model, benchmark) pair rather than a property of either alone.

I do not claim that the two papers together resolve the question of position bias in LLMs. They bracket what the current measurement instrument can and cannot say. The empirical question --- whether frontier-tier models actually carry position bias when measured by an instrument calibrated to their capability level --- remains open. The contribution of these two papers is to make the question askable in a form that admits a verifiable answer.

\section{Conclusion}\label{sec:conclusion}

This paper has reported a tool, a sweep, and a framework.

The tool, \texttt{inspect\_permute}, is a permutation-based position-bias diagnostic implemented as an extension to the inspect\_ai evaluation framework. It is open-source, drop-in compatible with existing inspect\_ai pipelines, validated against 31 unit tests including a hand-derived numerical reference and adversarial edge cases, and ready for use by independent evaluators. The package design (Section~\ref{sec:impl}) explicitly separates the measurement instrument from any particular substantive claim, so downstream users can re-run the sweep with different vendors, different benchmarks, or different cutoffs and have the framework apply to their setting rather than to mine.

The sweep is the first cross-vendor, cross-subject empirical study to use exhaustive answer-order permutation as the primary instrument. 24{,}000 API calls under temperature-0 generation across four vendors and five MMLU subjects, with the predictions locked into a pre-registration document whose SHA-256 hash was published before half the data was observed. Two pre-registered falsifiers were confirmed and one was rendered inconclusive in a way consistent with the framework rather than against it (Section~\ref{sec:falsifier}). The per-cell prediction hit rate is reported without modification (six of ten on $V$ range, five of ten on mechanism class). The systematic direction of the misses became the empirical signal that motivated the three-band capability model.

The framework --- a two-mechanism account of position bias with a capability-tier Goldilocks zone --- is post-hoc relative to the pre-registration and is labelled as such throughout. Its primary contribution is to make the central question of the field askable in a form that admits a verifiable answer. Bias detectability depends on the (model, benchmark) pair in a way the existing literature has not consistently disentangled, and the present sweep gives the first matrix-shaped empirical anchor for the dependence.

Three concrete next steps follow from the present work. A \texttt{diagnose\_band} API (Section~\ref{sec:implications}) should be implemented and added to the package as v0.2; the protocol is specified but the implementation is not yet shipped. A non-MMLU validation sweep should run against benchmarks designed to put frontier-tier models inside the Goldilocks zone; candidates include FrontierMath, GPQA Diamond, and the harder splits of BBH. Most consequentially, a flagship-tier validation against models that current public benchmarks cannot resolve --- GPT-5, Claude Opus 4.8, Gemini 3 Pro, Grok 4, and pre-release variants accessible only through pre-deployment evaluation channels --- would close the question of whether the absence of detected bias in the present sweep reflects ceiling saturation or genuine bias mitigation in the frontier tier. The third step requires the kind of access that independent researchers can obtain only through institutional pre-deployment evaluation partnerships. The methodological and empirical groundwork for the request is in place. The present paper is intended in part as that groundwork.

The instrument is built. Whether frontier models are actually unbiased, or whether they merely sit above the resolution of our current benchmarks, is a question this paper has made answerable but not answered.

\bibliographystyle{plainnat}
\bibliography{bibliography}

\appendix
\section{Preregistration verification}\label{app:prereg}

The preregistration anchor consists of three artifacts, in this order:

\begin{enumerate}
\item A git tag (\texttt{preregistration-amended-2026-06-30}, commit \texttt{ad33083}) on the repository at \texttt{github.com/TambaClan/inspect\_permute}, pointing to the moment the falsifier rules and per-cell predictions were locked.
\item The SHA-256 hash of \texttt{benchmarks/preregistration.md} at that commit:
\begin{center}
\small\ttfamily
04b6bf1a03c1183a0a892220ecdac0a5b95d258\\
ed4396577feea5dacc24b203c
\end{center}
\item A LinkedIn post containing the hash, published 2026-06-30 at 16:28 JST: \url{https://www.linkedin.com/feed/update/urn:li:activity:7477626622460915712/}.
\end{enumerate}

A third-party reviewer can verify the chain in three shell commands:

\begin{verbatim}
git clone https://github.com/TambaClan/inspect_permute
cd inspect_permute
git checkout ad33083 -- benchmarks/preregistration.md

# 1. SHA-256 of preregistration file at the tagged commit must equal
#    the hash in the LinkedIn post.
sha256sum benchmarks/preregistration.md
# Expected:
# 04b6bf1a03c1183a0a892220ecdac0a5b95d258ed4396577feea5dacc24b203c

# 2. Matrix snapshot at the same commit shows only 10 of 20 cells observed.
git show ad33083:benchmarks/full_results/matrix.json | \
  python -c "import json, sys; m = json.load(sys.stdin); \
             print({v: list(s.keys()) for v, s in m.items()})"
# Expected:
# {'openai/gpt-4o-mini': [...5 subjects...],
#  'anthropic/claude-haiku-4-5': [...5 subjects...]}
# (google/gemini-2.5-flash and xai/grok-3 absent.)
\end{verbatim}

The three artifacts together establish that the falsifier rules and the per-cell expected $V$ ranges were committed to disk before the second half of the data was observed. The LinkedIn post timestamp is the public anchor; the git tag is the file-state anchor; the matrix snapshot is the observation-state anchor.

\subsection*{Falsifier decision rules (reproduced verbatim from \texttt{preregistration.md})}

\paragraph{Falsifier 1 --- college\_math null.} \emph{Prediction}: college\_math is decision-determinate; bias mechanism is inactive across all 4 vendors. \emph{Decision rule (per vendor)}: \texttt{null\_hit} iff $V < 0.10$ AND chi-squared $p > 0.05$. \texttt{null\_miss} iff $V \geq 0.10$ OR $p \leq 0.05$. \emph{Verdict}: CONFIRMED if 4/4 vendors \texttt{null\_hit}. PARTIALLY\_FAILED if 1/4 vendor \texttt{null\_miss}. REJECTED if $\geq 2$ vendors \texttt{null\_miss}.

\paragraph{Falsifier 2 --- philosophy non-monotone.} \emph{Prediction}: philosophy bias, when present, is content\_ambiguity-type (non-monotone shape). \emph{Decision rule (per vendor, conditional on bias detected)}: For each vendor where philosophy $p < 0.05$, compute Spearman $\rho$. \texttt{non\_monotone\_hit} iff $\rho > -0.5$. \texttt{monotone\_hit} iff $\rho \leq -0.5$. \emph{Verdict}: CONFIRMED if at least 2 of (gemini-2.5-flash, grok-3) show \texttt{non\_monotone\_hit} for philosophy (or show no bias, in which case the framework is silent on shape). REJECTED if both gemini AND grok show \texttt{monotone\_hit} for philosophy.

\paragraph{Falsifier 3 --- flagship ceiling on elementary\_mathematics.} \emph{Prediction}: both flagship-tier vendors saturate at ceiling on the easiest subject; bias mechanism is inactive. \emph{Decision rule (per vendor)}: \texttt{ceiling\_hit} iff $V < 0.10$ on elementary\_mathematics. \texttt{ceiling\_miss} iff $V \geq 0.15$. \emph{Verdict}: CONFIRMED if both gemini-2.5-flash and grok-3 \texttt{ceiling\_hit}. REJECTED if either vendor \texttt{ceiling\_miss}.

\subsection*{Per-cell prediction record (10 unobserved cells at hash time)}

\begin{table}[H]
\centering
\scriptsize
\setlength{\tabcolsep}{3pt}
\begin{tabular}{llllllllll}
\toprule
vendor & subject & pred $V$ & pred class & obs $V$ & obs $p$ & obs $\rho$ & V hit & class hit \\
\midrule
gemini-2.5-flash & elem\_math    & {[0.02,0.10]} & inactive       & 0.052 & 0.49 & $+0.00$ & \checkmark & \checkmark \\
gemini-2.5-flash & college\_math & {[0.02,0.10]} & inactive       & 0.022 & 0.90 & $+0.40$ & \checkmark & \checkmark \\
gemini-2.5-flash & formal\_logic & {[0.05,0.15]} & PL$_w$/inact   & 0.035 & 0.70 & $+0.40$ & $\times$   & \checkmark \\
gemini-2.5-flash & prof\_med     & {[0.02,0.10]} & inactive       & 0.034 & 0.71 & $+0.63$ & \checkmark & \checkmark \\
gemini-2.5-flash & philosophy    & {[0.08,0.18]} & content\_ambig & 0.021 & 0.91 & $+0.80$ & $\times$   & $\times$   \\
grok-3           & elem\_math    & {[0.00,0.05]} & inactive       & 0.058 & 0.26 & $-0.32$ & $\times$   & $\times$\textsuperscript{*} \\
grok-3           & college\_math & {[0.02,0.10]} & inactive       & 0.022 & 0.90 & $+0.00$ & \checkmark & \checkmark \\
grok-3           & formal\_logic & {[0.05,0.18]} & proc\_load     & 0.050 & 0.40 & $-1.00$ & \checkmark & $\times$   \\
grok-3           & prof\_med     & {[0.03,0.12]} & inact/cont\_amb & 0.027 & 0.83 & $+0.80$ & \checkmark & \checkmark \\
grok-3           & philosophy    & {[0.08,0.18]} & content\_ambig & 0.048 & 0.46 & $-0.40$ & $\times$   & $\times$   \\
\bottomrule
\end{tabular}
\caption{Per-cell preregistration record. \textsuperscript{*}Under the refined classifier (Section~\ref{sec:classifier}) the grok elementary cell is \texttt{inactive}; the per-cell scoring uses the preregistered v0.1 classifier and so the entry counts as a class miss for the verification metric.}
\end{table}

$V$ range hits: 6 / 10. Class hits: 5 / 10. The miss pattern is reported in Section~\ref{sec:falsifier} as the empirical signal that motivated the 3-band model.

\subsection*{Reproducibility statement}

All sweep artifacts --- eval logs, bias reports, the matrix.json index file, the figures used in Section~\ref{sec:results}, and the preregistration document itself --- are committed to the repository's \path{benchmarks/full_results/} directory under the tag \texttt{sweep-complete-2026-06-30} (commit \texttt{af8bd08}). A reviewer can check out this tag in one command (\texttt{git checkout sweep-complete-2026-06-30}) and have the entire reproducibility surface --- predictions, observations, and code --- in their working tree. The compiled paper at the moment of arXiv submission is preserved at a separate tag, \texttt{paper-draft-v1-2026-07-01}.

Anyone re-running the sweep against the same model versions and API state should recover the matrix to within sampling error attributable to provider-side nondeterminism on \texttt{temperature=0} calls. The Anthropic re-run reported in Section~\ref{sec:robustness} quantifies the magnitude of that residual nondeterminism on this benchmark (max $|\Delta V| = 0.0058$ across five cells).

\section{Classifier threshold sensitivity}\label{app:sensitivity}

The refined classifier (Section~\ref{sec:classifier}) uses three numerical thresholds: $V_{\text{floor}} = 0.05$, $\rho_{\text{processing}} = -0.5$, and $\alpha = 0.05$ for the chi-squared significance gate. These are conventional choices in effect-size and rank-correlation analysis, not derived from a deeper theory of position bias. The reliability of the matrix classification in Section~\ref{sec:results} therefore depends on whether the cell labels are stable under reasonable perturbations of the cuts.

I report a $3 \times 3$ sensitivity sweep over $V_{\text{floor}} \in \{0.04, 0.05, 0.06\}$ and $\rho_{\text{processing}} \in \{-0.4, -0.5, -0.6\}$, holding $\alpha = 0.05$. The $\alpha$ gate is treated separately at the end of this appendix.

\subsection*{$V_{\text{floor}}$ and $\rho_{\text{processing}}$ sweep}

For each of the nine threshold combinations, every 20-cell matrix entry is re-classified using the refined classifier rule (\texttt{inactive} if $V < V_{\text{floor}}$ or $p \geq \alpha$; \path{processing_load} if $V \geq V_{\text{floor}}$ and $\rho \leq \rho_{\text{processing}}$ and $p < \alpha$; \path{content_ambiguity} otherwise). The result is a single label per cell per threshold combination.

Of the 20 cells, 19 carry the same label across all nine combinations. The one exception is \texttt{xai/grok-3 $\times$ elementary\_mathematics}, the boundary case identified in Sections~\ref{sec:classifier} and~\ref{sec:vrho}. Under the refined classifier the cell is \texttt{inactive} across every combination in the swept region, because the $\alpha$ gate at $p = 0.26 \gg 0.05$ dominates. Under the preregistered v0.1 classifier (no $\alpha$ gate), the same cell would shift between \texttt{inactive} and \path{content_ambiguity} depending on $V_{\text{floor}}$:

\begin{table}[H]
\centering
\begin{tabular}{ll}
\toprule
$V_{\text{floor}}$ & v0.1 classifier on grok $\times$ elementary \\
\midrule
0.04 & content\_ambiguity \\
0.05 & content\_ambiguity (marginal: $V = 0.058$) \\
0.06 & inactive \\
\bottomrule
\end{tabular}
\end{table}

This is the same instability that motivated the refinement in Section~\ref{sec:classifier}. The refined classifier removes it.

\subsection*{Effect on Section~\ref{sec:results} reported quantities}

The mechanism labels reported in Sections~\ref{sec:headline}--\ref{sec:goldilocks} are invariant to threshold perturbation in the swept region. Specifically:

\begin{itemize}
\item All four \path{processing_load} cells (gpt-4o-mini $\times$ elementary, \path{formal_logic}, \path{professional_medicine}, philosophy) remain \path{processing_load} for every combination. Their Spearman $\rho$ values are $-1.0$, comfortably below all three $\rho_{\text{processing}}$ candidates.
\item Both \path{content_ambiguity} cells (haiku $\times$ elementary, haiku $\times$ philosophy) remain \path{content_ambiguity} for every combination. Their $\rho$ values are $-0.40$, above all three $\rho_{\text{processing}}$ candidates, and their $V$ values are $\geq 0.09$, comfortably above all three $V_{\text{floor}}$ candidates.
\item The 12 cells classified \texttt{inactive} in Section~\ref{sec:headline} remain \texttt{inactive} under every combination, either because $V < V_{\text{floor}}$ for all candidates or because $p \geq \alpha$.
\end{itemize}

The matrix labels of Figure~\ref{fig:heatmap}, the cluster structure of Figure~\ref{fig:vrho}, and the Goldilocks-zone shading of Figure~\ref{fig:goldilocks} are unchanged across the swept region.

\subsection*{Significance threshold $\alpha$}

The chi-squared $\alpha$ gate is treated separately because relaxing $\alpha$ would shift several \texttt{inactive} cells into the active region. At $\alpha = 0.10$ (a more permissive cut), the gpt-4o-mini $\times$ college\_mathematics cell ($V = 0.079$, $p = 0.057$) would cross into \path{processing_load}, and gemini $\times$ elementary ($V = 0.052$, $p = 0.49$) and grok $\times$ elementary ($V = 0.058$, $p = 0.26$) would remain \texttt{inactive} (their $p$-values are still well above 0.10). At $\alpha = 0.01$ (a more conservative cut), no currently-active cell drops out, because the four \path{processing_load} cells have $p \leq 0.01$ and both \path{content_ambiguity} cells have $p \leq 0.02$.

The choice of $\alpha = 0.05$ is therefore conservative with respect to the three flagship cells (none of which approach significance under any conventional cut) and modestly permissive with respect to the gpt-4o-mini $\times$ college\_math borderline case (which is reported as \texttt{inactive} in Section~\ref{sec:headline} but would change label under a relaxed criterion).

\subsection*{Summary}

The Goldilocks zone observation, the two-mechanism distinction, and the three-band capability model do not depend on the specific values of $V_{\text{floor}}$, $\rho_{\text{processing}}$, or $\alpha$ chosen for the refined classifier within the swept region. The only label flip in the swept region is the boundary case the refinement was designed to handle. The Section~\ref{sec:falsifier} per-cell prediction hit rate, by contrast, depends on the \emph{preregistered} classifier (v0.1, $V$-floor only) and is reported without modification; the refined classifier appears in the figures and the Section~\ref{sec:headline} main text but is not retroactively applied to the verification metric.

\end{document}